\documentclass[sigplan,screen]{acmart}

\AtBeginDocument{%
  \providecommand\BibTeX{{%
    \normalfont B\kern-0.5em{\scshape i\kern-0.25em b}\kern-0.8em\TeX}}}

\usepackage[normalem]{ulem}
\usepackage{multirow}
\graphicspath{ {samples/figs/} }

\setcopyright{acmcopyright}
\copyrightyear{2021}
\acmYear{2021}
\acmDOI{10.1145/1122445.1122456}

\def\argmin{\mathop{\rm arg\,min}}

\begin{document}

\title{Deep Learning Based Page Creation for Improving E-Commerce Organic Search Traffic}


\author{Cheng Jie, Da Xu, Zigeng Wang, Wei Shen}
\affiliation{%
  \institution{Walmart Labs}
  \city{Sunnyvale}
  \state{California}
  \country{USA}
}
\email{{cheng.jie, da.xu, zigeng.wang0, wei.shen}@walmart.com}





\renewcommand{\shortauthors}{Cheng Jie and Da Xu, et al.}

\begin{abstract}
Organic search comprises a large portion of the total traffic for e-commerce companies. One approach to expand company's exposure on organic search channel lies on creating landing pages having broader coverage on customer intentions. In this paper, we present a transformer language model based organic channel page management system aiming at increasing prominence of the company's overall clicks on the channel. Our system successfully handles the creation and deployment process of millions of new landing pages. We show and discuss the real-world performances of state-of-the-art language representation learning method, and reveal how we find them as the production-optimal solutions.
\end{abstract}


\maketitle

\section{Introduction}
Since the prevalence of search engines, promoting marketing message on the search engine platform become vital in the success of E-Commerce business. There are mainly two sources of traffic in the search engine: paid search channel and organic search channel. Design strategies to expand company's exposure through organic channel is appealing to the industry because of its non-cost natural. Meanwhile, various independent findings demonstrate that on average, around 30$\%$ to 40$\%$ of E-Commerce total clicks come from organic search \cite{zhu2021time}. Together with the huge query volume on the search engine, there is still a great potential for E-Commerce sellers to boost their business on the unpaid organic channel.

The key principle of strategies on raising traffic at organic search channel is to improve the relevance between search queries and web pages under the rule of search ranking algorithm. According to the page ranking mechanism shared by Google \cite{Carrire12US, 6463486, Jie2018StochasticOI, PrashanthL2016CumulativePT}, we can infer three approaches leading to expanding the exposures of company's web page: Creating new web pages whose content are more related to the search queries, providing more direct connections between web pages loaded onto the search-engine offered sitemaps, and regularly updating web pages' content to keep in track of recent trending search queries. 

Over the years, a large body of literature studies the topic of search engine optimization(SEO), which is devoted to websites re-indexing and layout optimization. For instance, \cite{Zhang2017SearchEO, Yalcin2010WhatIS} studies the impact of link building and social sharing on increasing the rank of given websites, while \cite{Killoran2013HowTU, Bhandari2018ImpactOS, Dou2010BrandPS, LIN20181} propose methods of adding keywords and tags as website's metadata in order to enhance the relevance between search queries and context of website. Additionally, some AB testing and causal inference analysis frameworks \cite{Dover2011SearchEO, pinterest} are being proposed to leverage online experiments to select best web design and layout in optimizing organic traffic.  

Based on our industry experience, the traditional SEO methods focusing on enhancing the relevance of existing web pages still fall short of outstanding on many generic queries search results. We observed that item pages are too specific to match queries such as ``nike shoes new design 2021'' and ``covid test kits''. Although company do have certain non-item specific category webpages used to capture broader purchase intentions, they are too static to follow the evolving trend of search queries. As a result, a system of generating new web pages capturing generic search queries proves to have a great potential. 

For 
Since there is a limited quota in search engine webmaster, the repo onto which customer upload its pages for crawling and indexing, new page creation process must be subject to efficient management. Maximizing potential customer intention coverage through page generation amounts to finding only the customer intentions haven't yet been covered by the existing websites. Moreover, for each batch of newly created topic pages, we also need to remove the redundant ones in order to optimally utilize the quota. 

\subsection*{Two stage e-commerce topic page creation system}
In this paper, we design a two stage web page creation system dedicated to increase the clicks and conversion of company's products on the unpaid channel of search engine. The solution proposed in our paper is currently in production for the multi-million-scale web pages management for Walmart's e-commerce business. On the high level, our page creation system is centered around generating ``topic keywords'' websites so that we first select a keyword phrase indicating a certain set of purchase intention, and then build a web page of products relevant to the keyword. As demonstrated in detail later in section \ref{section:system_overview}, the work stream of our system comprises of two critical components:
\begin{itemize}
    \item a topic keywords clustering algorithm based on the keywords' vector representations of their indicated purchase intention
    \item a topic keywords deduplication algorithm which filter out candidate keywords having duplicated customer intention from existing web pages;
\end{itemize}
Toward our goals, we first design a transformer-based query embedding model to extract vector representations of search queries, which captures the customer purchase intention integrated in each query. As illustrated in the section \ref{section:query-embedding}, geometric characteristics of purchase intention embedding model are unlike the normal embedding model pre-trained from popular public modules. As detailed in section \ref{section:clustering-deduplication}, both the keywords clustering and deduplication services are built upon extending the query customer intention embedding results. Note that existing types of web pages provided on the search engine sitemap are not limited to topic pages, making both two services necessary. 

We thoroughly examine the performance of the proposed topic page generation process in Section \ref{section: experiment}. As expected, topic page creation processes driven by deep learning language model puts clear advantage in capture more clicks and conversions.

\section{Search Engine Optimization: Topic page creation system overview}
\label{section:system_overview}
\begin{figure*}[hbt]
    \centering
    \includegraphics[width=\linewidth]{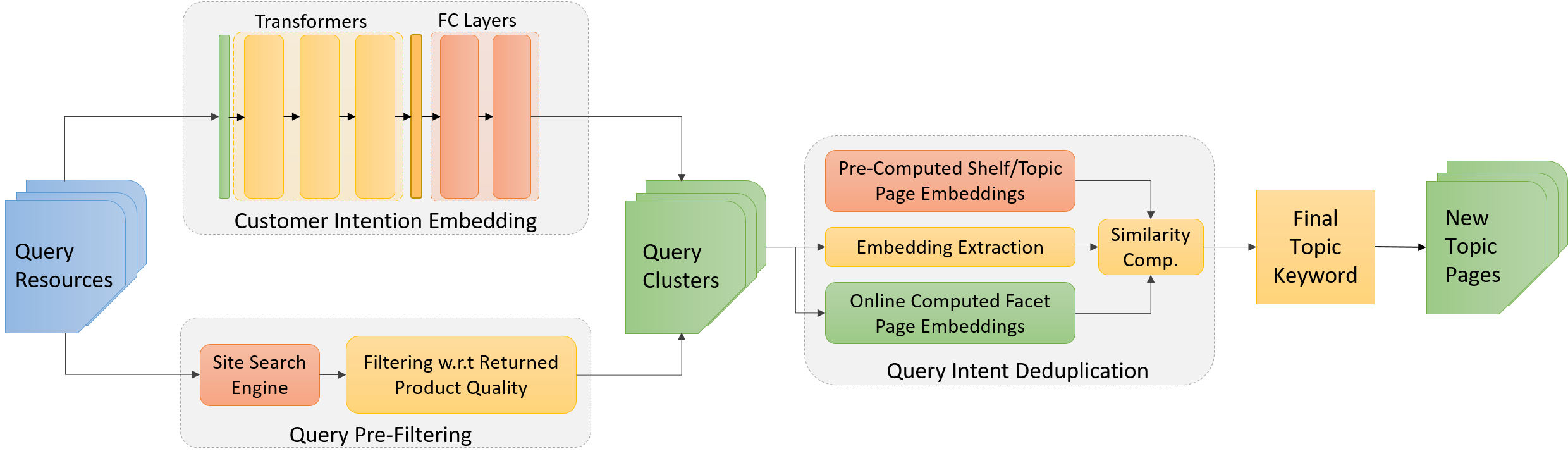}
    \caption{Overview of the infrastructure topic generation system}
    \label{fig:seo_generation_whole_sys_comprehensive}
\end{figure*}

In this section, we present an end-to-end topic page creation system currently being deployed in production on a regular basis. As displayed in the chart \ref{fig:seo_generation_whole_sys_comprehensive}, candidate queries which could potentially become topic keywords are selected from multiple resources including internal search logs, query-ad search engine marketing(SEM) report from search engines and recent general trending queries from social media, etc. A query customer intention embedding model, which seeks learning vector representation of query's purchase intention, is developed to assist the query clustering and de-duplication process. With the query customer intention model, we leverage its functionality and extend it on the task of query clustering and query de-duplication. 

In what follows, candidate queries extracted from multiple sources will first go through an initial filtering process to clean up offensive and non-product related negative queries. For the queries left after filtering, a clustering and a de-duplication algorithms will be applied consecutively to remove duplicate customer intention coverage, from within the set of input queries and against existing web pages, respectively. The eventual set of queries after all the filtering and de-duplicating processes are accepted as topic keywords, based on which topic pages are generated. Here, for each keyword, we use internal organic search API to collect top items related to the keyword, and create a topic page according to the returned search items.



\section{Query customer intention embedding}
\label{section:query-embedding}
Customer intention of a query represents the set of product related landing pages that leads to a click action after the query is triggered on the search engines. As opposed to general query embedding model trained on context ranging from various categories, customer intention embedding is built as an numerical embodiment of only the commercial/product sense of queries. For example, the phrase ``banana republic'' in general can refer either a political term or a fashion collection, but our query customer intention embedding model only represents its meaning as a production brand. By the definition of query customer intention, it is easy to see that if two queries share a large portion of clicked items, their intentions should be close to each other. Therefore, customer intention model is designed to reflect the co-click relationships among search queries. In this section, we walk through the design and the training process query customer intention model, a parallel work on customer intention of landing pages can be found in \cite{Jie2021BiddingVC, Jie2018DecisionMU, ZHAO2018619, huang2022unfolding}.

\textbf{Interactive metric}. We propose a measure named as interactive metric(I) in order to calibrate the extent of similarity between customer intentions of two queries. Formally, given two queries Q1 and Q2, we first find the numbers of clicks of the two ads on their co-clicked and denote them as $CLK_{(Q1coQ2)}$ and $CLK_{(Q2coQ1)}$. With the numbers of total historical clicks of the two ads $CLK_{Q1}$ and $CLK_{Q2}$, the I value between Q1 and Q2 is defined as 
\begin{align}\label{eq:interactive-metric}
I_{Q1, Q2} = \sqrt{\frac{CLK_{(Q1coQ2)} * CLK_{(Q2coQ1)}}{CLK_{Q1} * CLK_{Q2}}},
\end{align}

\begin{figure}[h]
  \centering
  \includegraphics[width=\linewidth]{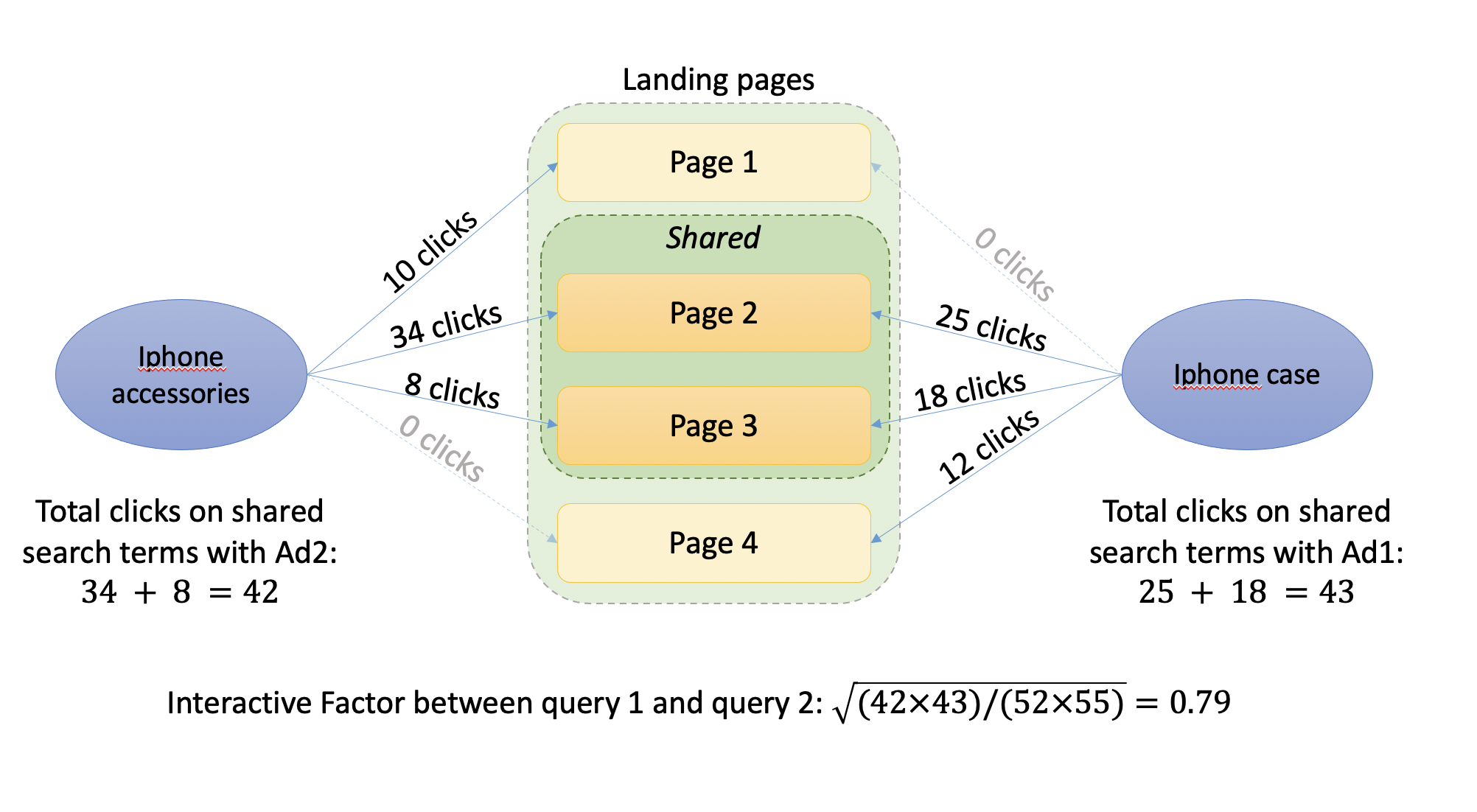}
  \caption{Interactive metric: an example}
   \label{fig:interactive_factor}
\end{figure}

Fig \ref{fig:interactive_factor} provide a real-world example to further demonstrate how interactive factor of two SEM ads is calculated. In fig \ref{fig:interactive_factor}, query ``iphone accessories'' and ``iphone case'' have shared clicked landing pages ``Page 2'' and ``Page 3'', along with corresponding 42 and 43 clicks on the shared queries. Moreover, the total number of historical clicks for ``iphone accessories'' and ``iphone case'' are 52 and 55 respectively, leading to the interactive value between the two ads be $\sqrt{(42 * 43)/(52 * 55)} = 0.798$.

\textbf{search query tokenization}.
Given a query $Q$, its word tokens vector is the main part of its quantified features for model input. Additionally, two similar queries with different facets account for two different customer intentions by our topic generation criteria, with facets refer to color, gender, product type etc. In order to highlight query's implied facets, we leverage the company established facets understanding API to extracts facets information of each query. As a result, the eventual query token vector $T_Q$ is the concatenation of query's raw word tokens and the token of its extracted facets. 

\textbf{Query customer intention representation model}.
Word embedding techniques have become ubiquitous since the publication of word2vec\cite{Mikolov2013DistributedRO}. Starting from 2 layer embedding neural network which assign each word an embedding vector, various forms of word embedding model have been established for different objectives. In particular, sequential language models applied on decode-encode task has drawn the most attention from academia and industry research\cite{Bahdanau2015NeuralMT}. Among different kinds of sequential models, self-attention driven transformer has emerged as an optimal architecture for sentence encode-decode tasks such as translation, query matching and classification. The major advantage of transformer over other RNN-based models is its ability to avoid long dependency issues. Well-known transformer based architectures include BERT\cite{Devlin2019BERTPO} and USE\cite{Cer2018UniversalSE}. 

\begin{figure}[h]
  \centering
  \includegraphics[width=\linewidth]{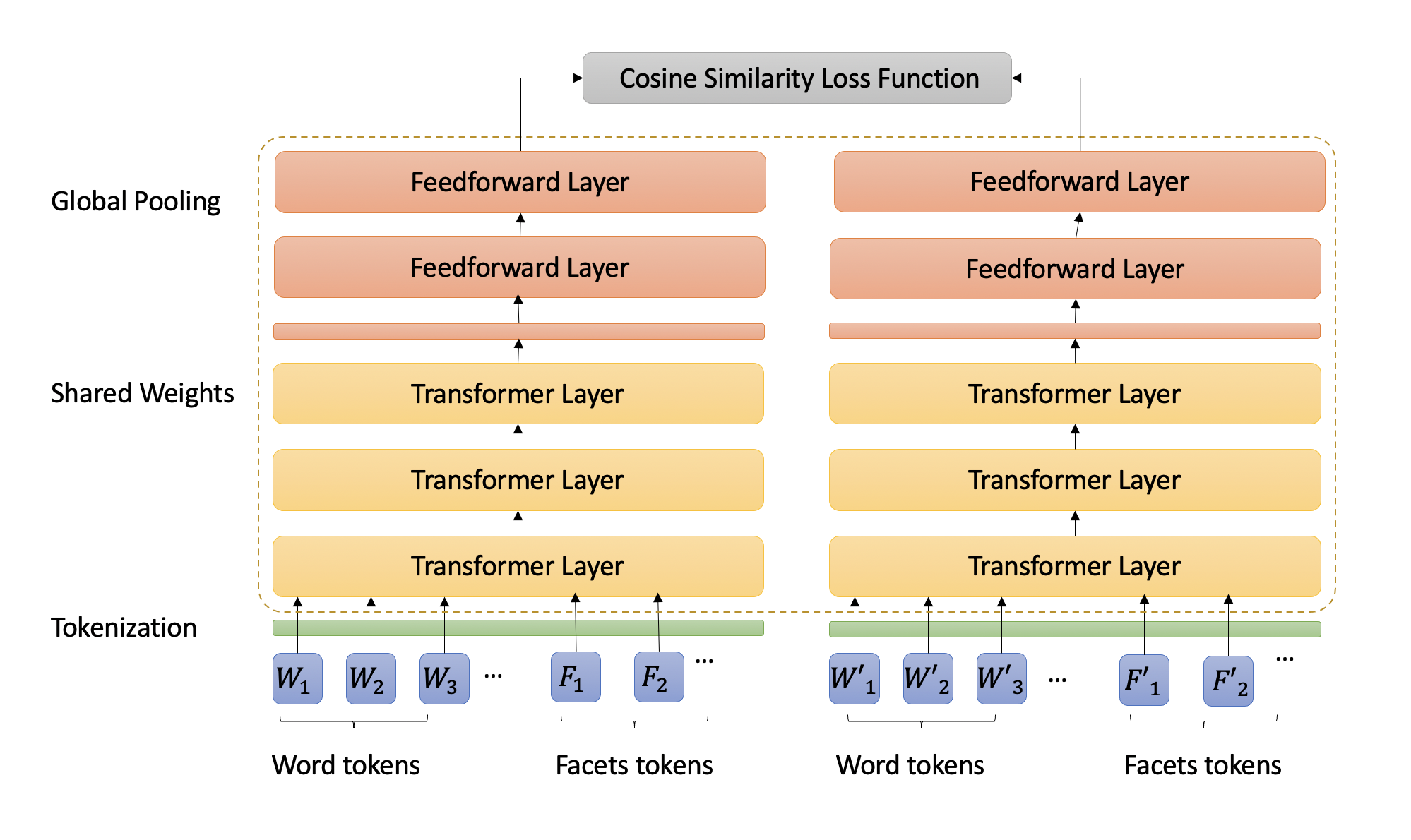}
  \caption{Query intention embedding model}
   \label{fig:intention_representation}
\end{figure}

In light of the capability of transformer architecture, we built a transformer driven query customer intention embedding model.
As we show in Fig \ref{fig:intention_representation}, for a given query $Q$ and its tokenized feature $T_Q$, the model will consecutively go through an initial embedding layer, $3$ transformer layers, a dense pooling layer, and two feed-forward layers before generating the final $512$-dimension \emph{normalized} output vector.

\textbf{Training data}.
Similar to \cite{Jie2021BiddingVC}, the data we use for training the representation learning model is the $\texttt{search\_term\_report}$ from search engine, which provides the historical statistics of interactions (e.g. clicks, impressions) between queries and their click-relevant landing pages. Specifically, for each query, we will extract historical click numbers between the query and landing pages that leads to the clicks. 
Together with interactive metric $I$ defined at \ref{eq:interactive-metric}, we create a data-set $\mathcal{D}$ containing all the tuples of queries having co-clicked together with their interactive metric. In order to cover larger support of the distribution, we create $-1$ interactive metric negative tuples of queries in our dataset just as \cite{Jie2021BiddingVC}. The best practice of ratio between positive tuples and negative tuples should approximately equal to the average positive interactive metric in the data.

\textbf{Model training}.
Let $f_\theta(\cdot)$ denote a customer intention model
with parameter vector $\theta$. Given a query tuple $(Q_i, Q_j)$ along with their interactive metric $I_{ij}$, we define the loss function as
\begin{align}
-I_{ij}\log\sigma\big(f_\theta(T_{Q_i})^Tf_\theta(T_{Q_j})\big),
\end{align}
where $\sigma(\cdot)$ is the sigmoid function. The inner product of $f_\theta(T_{Q_i})^Tf_\theta(T_{Q_j})$ captures the cosine similarity between the embeddings of $(Q_i, Q_j)$, given that output vectors of the model $f_\theta(\cdot)$ are normalized. 
The optimization problem for finding the optimal $\theta $ is now given by: 
\begin{align}
\theta^{\star} = 
\argmin_{\theta \in \Theta}
\sum_{(Q_i, Q_j) \in DT} -I_{ij}\log\sigma(f_\theta(Q_i)^Tf_\theta(Q_j)), 
\label{eq:model-objective}
\end{align}
The objective (\ref{eq:model-objective}) indicates that the larger the interactive metric between two queries, the more impact this query instance will bring when determining model parameter $\theta$. The structure of the model, together with the procedure for calculating the loss function, are presented in Figure \ref{fig:intention_representation}.

\section{Topic keywords clustering and de-duplication algorithms}
\label{section:clustering-deduplication}
As mentioned earlier, clustering and de-duplication algorithms are designed to prune the input queries so that the eventual topic keywords represent new and mutually exclusive customer intentions. Therefore, customer intention embedding model on the queries play an essential role on the two tasks. In this section, we illustrate each step of query clustering and de-duplication process.

\subsection{Topic keywords clustering}
queries clustering algorithm is designed to identify duplicated customer intentions within the batch of input candidate topic keywords. Due to the high volume of queries, though many efficient machine learning models have been introduced, it is still impractical to apply the clustering algorithms that require computing all the pair-wise distances \cite{zhu2022clustering, zhu2021news}. Here, we present a multi-stage method that leverages the company's predefined taxonomy and significantly reduces the computational cost.

\textbf{Query classification}. The first step of the multi-stage clustering algorithm is to classify each query into one of the \emph{product types}, which can be any taxonomy that is labeled for the items: electronics, beverage, etc. Most companies have the predefined taxonomy for each item, which should be actively exploited. Here, we first retrieve customer intention embedding vectors for all product type and queries. And for each query, we classify it to the product type having the closest embedding vector to the its embedding vector in terms of cosine similarity. 
 
\textbf{Clustering within each product type}.
Following the classification, we apply the "bottom-up" Agglomerative clustering using embedding vectors as features to create mutually exclusive query groups within each product type. Naturally, the cosine distance is employed as the \emph{linkage metric}, and it also allows us to determine the threshold based on which the final clusters are formed. We point out that the first classification step significantly reduces the computation complexity compared with directly clustering all the input queries. 

\subsection{Query deduplication}
\begin{figure}[h]
  \centering
  \includegraphics[width=\linewidth]{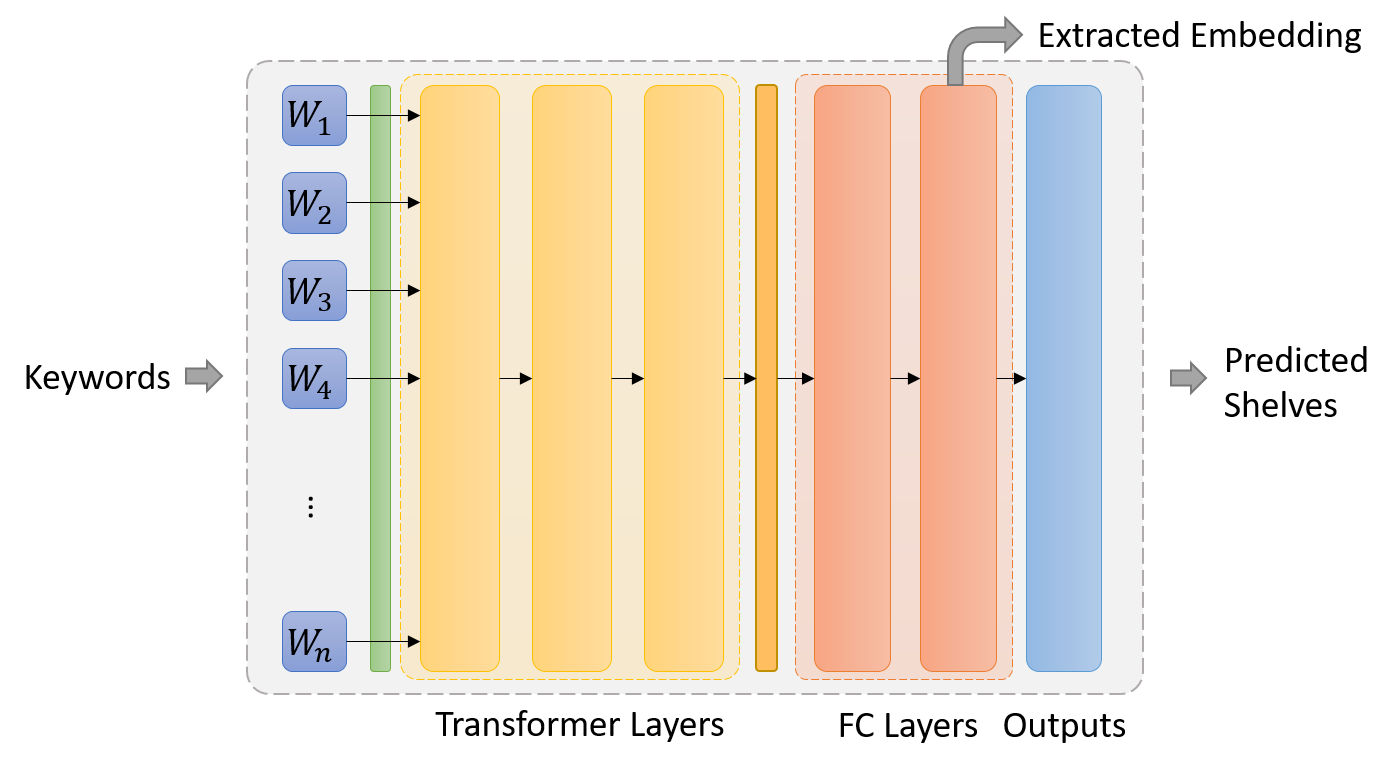}
  \caption{Keyword classification and embedding extraction}
   \label{fig:keyword_classification}
\end{figure}

Query deduplication algorithm is designed to filter out the duplicated candidate queries which cover close intents of existing category web pages. The web pages vary in different page types from shelf to facet pages. The total number of existing category pages is too large to make direct page classification as a feasible solution. Leveraging query intention embeddings provided in section \ref{section:query-embedding}, we transform the page deduplication problem into a query-page semantic matching problem, in which we filter queries having semantically close existing pages with regard to embedding vectors. We layout the solution with the following two stages:

\textbf{Deduplication task-specific query embedding}. 
In order to better serve our query-deduplication service, 
we apply transfer learning method to extend the query embedding model trained at section \ref{section:query-embedding} on our task-specific data \cite{li2021frequentnet}.  Specifically, we design a supervised classification model predicting the closest web category pages for each query. As shown in Fig ~\ref{fig:keyword_classification}, the classification model is built upon query intention embedding model introduced at Fig ~\ref{fig:intention_representation} with an additional classification layer indicating predicted web category pages. We extract subset of query-category web pages interactive data from $\texttt{search\_term\_report}$ as the labeled dataset, and utilize cross-entropy as the objective function for parameter tuning. The parameters retrieved from the embedding model at section \ref{section:query-embedding} are serving as the starting point for the model training. The second to last layer of the trained classification model are extracted as the deduplication task-specific embeddings for both candidate queries and web page titles, in that the hidden representations used to distinguish shelves can be diverse enough to describe the relations among different user queries.

\textbf{Query deduplication}. In order to efficiently deduplicate queries against existing web pages with various types, we design the following query-page similarity computation framework explained at Fig \ref{fig:keyword_simi_analysis}. The framework is composed of two major parts built for deduplicating queries against different category pages, which is explained below: 
\begin{itemize}
    \item For shelf pages with relatively small quantity, we pre-compute the embedding vectors for all the existing ones and load them into the memory. As displayed on the upper part of Fig \ref{fig:keyword_simi_analysis}, for each new query, we compute its cosine similarities with all the existing shelf pages and store them for later use.  
    \begin{figure}[!h]
  \centering
  \includegraphics[width=\linewidth]{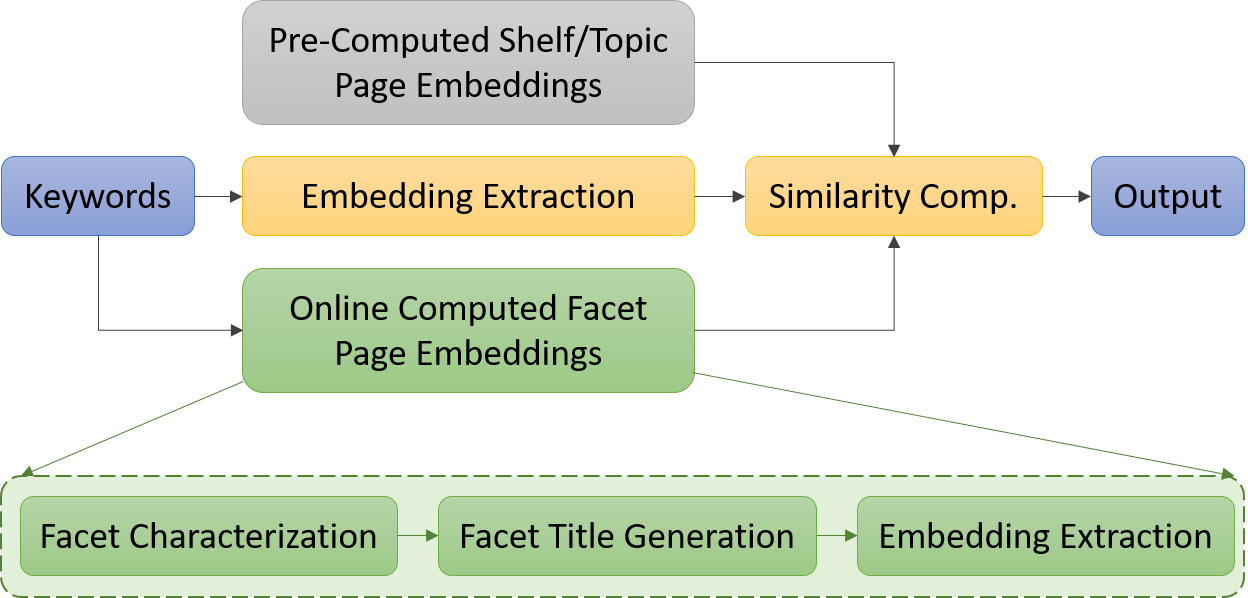}
  \caption{Keyword-Page similarity analysis}
   \label{fig:keyword_simi_analysis}
\end{figure}
    \item For facet pages with finer granularity than shelf pages, the total quantity is too high to pre-compute embedding vectors of all pages and store them in memory. As a result, we design an online query-facet deduplication approach described by the lower part of Fig \ref{fig:keyword_simi_analysis}. In specific, we leverage shelf page classification and n-gram matching techniques to extract facet information of the input query. Given the query's characterized facets, we narrow down the facet pages deduplication task to computing cosine similarities of the input query with only a handful of related facet pages, reducing our computational effort dramatically.    
\end{itemize}
The cosine similarity values computed from the above steps are used to filter out input queries with referred to a high threshold. In practice, we choose a value between 0.85 to 0.88 as the deduplication criterion. 

\section{Experiments and Analysis}
\label{section: experiment}
We conducted two online experiments to answer the following questions:

\textbf{Q1:} Does generating topic pages really help expanding the overall clicks of company's websites on the organic search channel? 

\textbf{Q2:} How does our clustering$\&$deduplication method compare with other baseline method?

To this end, we design two online experiments to investigate the performances of topic page generation.

For the two experiments, we are given roughly 5 million raw search queries from $\texttt{search\_term\_report}$ as the source of topic keywords generation. Note that according to the google webmaster tool, the quota of new web pages allowed to be uploaded is less than 5 million, justify the need of an optimal strategy in selecting subset of queries as topic keywords. We leverage our keywords clustering$\&$deduplication method to filter out roughly 4 million queries, and select 500K(approximately half of the quota) for the following experiments.  

\begin{figure}[h]
  \centering
  \includegraphics[width=\linewidth]{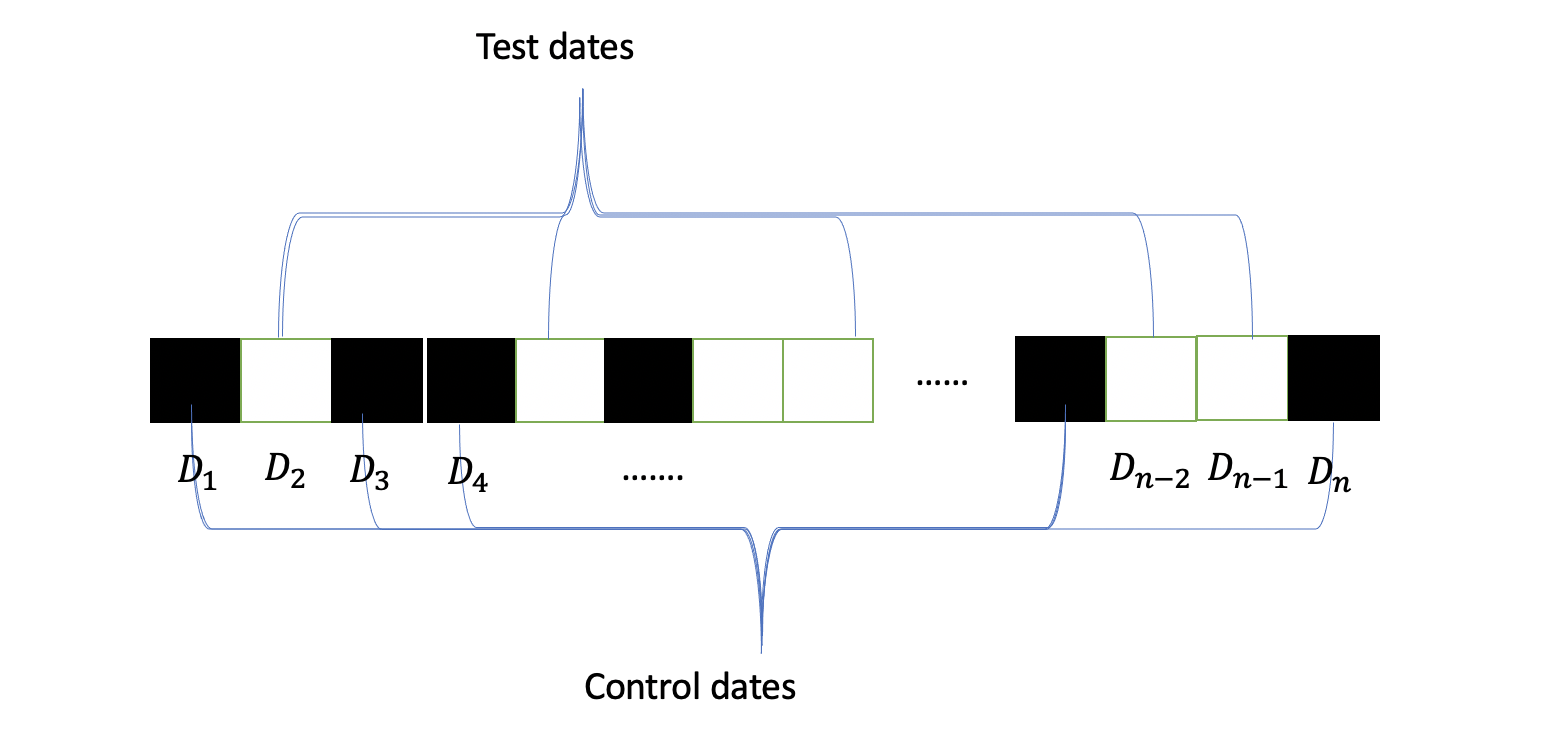}
  \caption{Test and Control dates separation}
   \label{fig:ab_test_periods}
\end{figure}

Although it cannot split traffic on two campaigns like Adwords, Google search console provides functionality to ``pause'' and ``unpause'' websites provided on the repo. Based on that functionality, we have the following high level experimental design: First, given a time window with certain number of days, we randomly and evenly split the time window into 2 sets of dates. We then select one set of dates as the control group and the other as the test group with corresponding treatments, and compare the performances of the two sets of dates after the test period. The control and test dates split method can be explained in the Fig~\ref{fig:ab_test_periods}.

\begin{figure}[h]
  \centering
  \includegraphics[width=\linewidth]{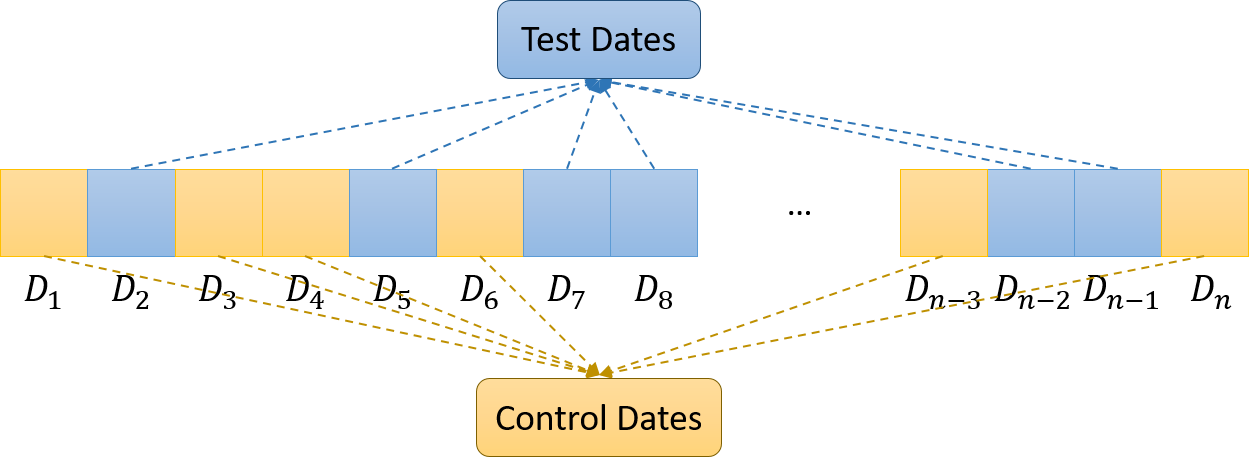}
  \caption{Test and Control dates separation}
   \label{fig:ab_test_periods2}
\end{figure}

\subsection*{Online experiment 1: Test whether generating topic pages can improve marketing exposure}
We design the experiment to test whether adding new topic pages on top of existing web pages can help expanding the overall clicks through organic search traffic. The new topic pages are first uploaded to the Google search console, and we first leave those pages active for two weeks in order to allow the search engine to fully crawl and store them into the database. Through power analysis, we select 120 days as the total test period, with the first 60 days as AA and the second 60 days as AB. For each of the AA and AB period, we use the same mechanism to split the 60 days into two evenly sets of dates, and let them as control and test group correspondingly. 

The AA period is used to justify whether dates splitting mechanism ensures a fair comparison. During the AA period, we only let original websites before topic creation to be active for 60 days, and check the clicks of the dates in control and test groups. For the AB period, the new topic pages are paused on control dates and active on the test dates. The final statistics for the control vs test group are summarized in Table~\ref{table:online-experiment-results-1}. According to the table, during the AA period, the clicks from the control and test dates are very close to each other, justifying the fairness of the dates split (with a two sample t statistics 0.1 and p-value 0.84). Meanwhile, the AB period reveals that new topic pages do have a positive impact on the overall clicks of company's web pages at organic search channel, the two sample t statistics equals 1.69 with p-value equals 0.048. 

\begin{table}
\caption{Online testing results: with topic pages vs without topic pages}
\begin{tabular}{cc|cc}
    \toprule
    Test  & Metric(Relative) & Control dates & Test dates\\ 
    \midrule
    AA period & Clicks & 100$\%$ & 101$\%$ \\ 
    AB period & Clicks & 100$\%$ & 104$\%$ \\ 
     \bottomrule
  \end{tabular}
\label{table:online-experiment-results-1}
\end{table}

\subsection*{Online experiment 2: Comparison against baseline algorithm}
Here, the baseline strategy for creating topic keywords is selecting queries based on highest clicks. Therefore, we first extract the top 500K queries with the highest clicks, and create corresponding web pages, and treat them as control group. Note that there are around 40$\%$ overlap between the control pages and test group pages generated through clustering$\&$deduplication algorithm. In order to save space, we only upload the topic pages of control group which do not belong to the test group. Similar to offline experiment 1, we first make sure all the pages are fully crawled and indexed.

We select another 120 days period and make the first 60 days as AA and second 60 days as AB, and split each of the two periods randomly and evenly into two a control and test set of dates under similar approaches in experiment 1. Only control group web pages are active during the AA period. For AB period, the web pages in the control group are active on the control set of dates while the web pages in the test group are active on the test set of dates. Notice that here, we only need to turn on and off the pages that are not overlapping in the two groups to reduce computational demand.  

The test result is summarized in Table~\ref{table:online-experiment-results-2}, where it exemplifies that test group topic pages created through clustering and 
deduplication algorithms capture more clicks than the control group topic pages, which is created solely based number of clicks. Follow the online experiment 1, we also conduct a two sample t statistics on the AB test result, and the t statistics equals to 3.01, with p value less than 0.01. Moreover, if we extracted comparisons of statistics only for the non-overlap web pages, the differences between control and test groups are even more significant, which is presented in Table~\ref{table:online-experiment-results-2-nonoverlap} (t statistics 6.04, p-value much less than 0.01). 

\begin{table}
\caption{Online testing results: Control topic pages vs test topic pages}
\begin{tabular}{cc|cc}
    \toprule
    Test  & Metric(Relative) & Control dates & Test dates\\ 
    \midrule
    AA period & Clicks & 100$\%$ & 99$\%$ \\ 
    AB period & Clicks & 100$\%$ & 111$\%$ \\ 
     \bottomrule
  \end{tabular}
\label{table:online-experiment-results-2}
\end{table}

\begin{table}
\caption{Online testing results: Control group vs test group for only non-overlap pages}
\begin{tabular}{cc|cc}
    \toprule
    Test  & Metric(Relative) & Control dates & Test dates\\ 
    \midrule
    AA period & Clicks & 100$\%$ & 101$\%$ \\ 
    AB period & Clicks & 100$\%$ & 124$\%$ \\ 
     \bottomrule
  \end{tabular}
\label{table:online-experiment-results-2-nonoverlap}
\end{table}

\section{Conclusion}
This paper introduces a two-step page creation system that integrates modern representation learning with the Transformer language model. We describe the detailed development infrastructure that may bring insights to both practitioners and researchers in this domain. The offline and online experiments show that the proposed system compares favorably to the alternatives in terms of expanding the company's overall exposures on the organic search channel. 
Our successful deployment for Walmart e-commerce further reveals application with modern representation learning as a scalable solution for industrial new page generation problem. 

\bibliographystyle{ACM-Reference-Format}
\bibliography{irs_workshop_arxiv}


\begin{thebibliography}{24}


\ifx \showCODEN    \undefined \def \showCODEN     #1{\unskip}     \fi
\ifx \showDOI      \undefined \def \showDOI       #1{#1}\fi
\ifx \showISBNx    \undefined \def \showISBNx     #1{\unskip}     \fi
\ifx \showISBNxiii \undefined \def \showISBNxiii  #1{\unskip}     \fi
\ifx \showISSN     \undefined \def \showISSN      #1{\unskip}     \fi
\ifx \showLCCN     \undefined \def \showLCCN      #1{\unskip}     \fi
\ifx \shownote     \undefined \def \shownote      #1{#1}          \fi
\ifx \showarticletitle \undefined \def \showarticletitle #1{#1}   \fi
\ifx \showURL      \undefined \def \showURL       {\relax}        \fi
\providecommand\bibfield[2]{#2}
\providecommand\bibinfo[2]{#2}
\providecommand\natexlab[1]{#1}
\providecommand\showeprint[2][]{arXiv:#2}

\bibitem[\protect\citeauthoryear{Bahdanau, Cho, and Bengio}{Bahdanau
  et~al\mbox{.}}{2015}]%
        {Bahdanau2015NeuralMT}
\bibfield{author}{\bibinfo{person}{Dzmitry Bahdanau},
  \bibinfo{person}{Kyunghyun Cho}, {and} \bibinfo{person}{Yoshua Bengio}.}
  \bibinfo{year}{2015}\natexlab{}.
\newblock \showarticletitle{Neural Machine Translation by Jointly Learning to
  Align and Translate}.
\newblock \bibinfo{journal}{\emph{CoRR}}  \bibinfo{volume}{abs/1409.0473}
  (\bibinfo{year}{2015}).
\newblock


\bibitem[\protect\citeauthoryear{Bhandari and Bansal}{Bhandari and
  Bansal}{2018}]%
        {Bhandari2018ImpactOS}
\bibfield{author}{\bibinfo{person}{R. Bhandari} {and} \bibinfo{person}{A.
  Bansal}.} \bibinfo{year}{2018}\natexlab{}.
\newblock \showarticletitle{Impact of Search Engine Optimization as a Marketing
  Tool}.
\newblock \bibinfo{journal}{\emph{Jindal Journal of Business Research}}
  \bibinfo{volume}{7} (\bibinfo{year}{2018}), \bibinfo{pages}{23 -- 36}.
\newblock


\bibitem[\protect\citeauthoryear{Carri{\`e}re}{Carri{\`e}re}{[n.d.]}]%
        {Carrire12US}
\bibfield{author}{\bibinfo{person}{S. Carri{\`e}re}.}
  \bibinfo{year}{[n.d.]}\natexlab{}.
\newblock \showarticletitle{(12) United States Patent Page (54) Method for Node
  Ranking in a Linked Database}.
\newblock


\bibitem[\protect\citeauthoryear{Cer, Yang, yi~Kong, Hua, Limtiaco, John,
  Constant, Guajardo-Cespedes, Yuan, Tar, Sung, Strope, and Kurzweil}{Cer
  et~al\mbox{.}}{2018}]%
        {Cer2018UniversalSE}
\bibfield{author}{\bibinfo{person}{Daniel~Matthew Cer}, \bibinfo{person}{Yinfei
  Yang}, \bibinfo{person}{Sheng yi Kong}, \bibinfo{person}{Nan Hua},
  \bibinfo{person}{Nicole Limtiaco}, \bibinfo{person}{Rhomni~St. John},
  \bibinfo{person}{Noah Constant}, \bibinfo{person}{Mario Guajardo-Cespedes},
  \bibinfo{person}{Steve Yuan}, \bibinfo{person}{C. Tar},
  \bibinfo{person}{Yun-Hsuan Sung}, \bibinfo{person}{B. Strope}, {and}
  \bibinfo{person}{R. Kurzweil}.} \bibinfo{year}{2018}\natexlab{}.
\newblock \showarticletitle{Universal Sentence Encoder}.
\newblock \bibinfo{journal}{\emph{ArXiv}}  \bibinfo{volume}{abs/1803.11175}
  (\bibinfo{year}{2018}).
\newblock


\bibitem[\protect\citeauthoryear{Devlin, Chang, Lee, and Toutanova}{Devlin
  et~al\mbox{.}}{2019}]%
        {Devlin2019BERTPO}
\bibfield{author}{\bibinfo{person}{J. Devlin}, \bibinfo{person}{Ming-Wei
  Chang}, \bibinfo{person}{Kenton Lee}, {and} \bibinfo{person}{Kristina
  Toutanova}.} \bibinfo{year}{2019}\natexlab{}.
\newblock \showarticletitle{BERT: Pre-training of Deep Bidirectional
  Transformers for Language Understanding}. In
  \bibinfo{booktitle}{\emph{NAACL-HLT}}.
\newblock


\bibitem[\protect\citeauthoryear{Dou, Lim, Su, Zhou, and Cui}{Dou
  et~al\mbox{.}}{2010}]%
        {Dou2010BrandPS}
\bibfield{author}{\bibinfo{person}{Wenyu Dou}, \bibinfo{person}{Kai~H. Lim},
  \bibinfo{person}{C. Su}, \bibinfo{person}{Nan Zhou}, {and}
  \bibinfo{person}{Nan Cui}.} \bibinfo{year}{2010}\natexlab{}.
\newblock \showarticletitle{Brand Positioning Strategy Using Search Engine
  Marketing}.
\newblock \bibinfo{journal}{\emph{MIS Q.}}  \bibinfo{volume}{34}
  (\bibinfo{year}{2010}), \bibinfo{pages}{261--279}.
\newblock


\bibitem[\protect\citeauthoryear{Dover and Dafforn}{Dover and Dafforn}{2011}]%
        {Dover2011SearchEO}
\bibfield{author}{\bibinfo{person}{Danny Dover} {and} \bibinfo{person}{Erik
  Dafforn}.} \bibinfo{year}{2011}\natexlab{}.
\newblock \showarticletitle{Search Engine Optimization (SEO) Secrets}.
\newblock


\bibitem[\protect\citeauthoryear{engineer}{engineer}{2015}]%
        {pinterest}
\bibfield{author}{\bibinfo{person}{Julie Ahn |~Pinterest engineer}.}
  \bibinfo{year}{2015}\natexlab{}.
\newblock \bibinfo{booktitle}{\emph{Demystify SEO with experiments}}.
\newblock
\urldef\tempurl%
\url{https://medium.com/pinterest-engineering/demystifying-seo-with-experiments-a183b325cf4c}
\showURL{%
\tempurl}


\bibitem[\protect\citeauthoryear{Huang, Zhao, Wang, Li, Yang, Feng, Xu, Zhu,
  and Chen}{Huang et~al\mbox{.}}{2022}]%
        {huang2022unfolding}
\bibfield{author}{\bibinfo{person}{Xiao Huang}, \bibinfo{person}{Yuhui Zhao},
  \bibinfo{person}{Siqin Wang}, \bibinfo{person}{Xiao Li}, \bibinfo{person}{Di
  Yang}, \bibinfo{person}{Yu Feng}, \bibinfo{person}{Yang Xu},
  \bibinfo{person}{Liao Zhu}, {and} \bibinfo{person}{Biyu Chen}.}
  \bibinfo{year}{2022}\natexlab{}.
\newblock \showarticletitle{Unfolding community homophily in US metropolitans
  via human mobility}.
\newblock \bibinfo{journal}{\emph{Cities}}  \bibinfo{volume}{129}
  (\bibinfo{year}{2022}), \bibinfo{pages}{103929}.
\newblock


\bibitem[\protect\citeauthoryear{Jie}{Jie}{2018}]%
        {Jie2018DecisionMU}
\bibfield{author}{\bibinfo{person}{Cheng Jie}.}
  \bibinfo{year}{2018}\natexlab{}.
\newblock \showarticletitle{Decision Making Under Uncertainty: New Models and
  Applications}.
\newblock


\bibitem[\protect\citeauthoryear{Jie, PrashanthL., Fu, Marcus, and
  Szepesvari}{Jie et~al\mbox{.}}{2018}]%
        {Jie2018StochasticOI}
\bibfield{author}{\bibinfo{person}{Cheng Jie}, \bibinfo{person}{A.
  PrashanthL.}, \bibinfo{person}{M. Fu}, \bibinfo{person}{S. Marcus}, {and}
  \bibinfo{person}{Csaba Szepesvari}.} \bibinfo{year}{2018}\natexlab{}.
\newblock \showarticletitle{Stochastic Optimization in a Cumulative Prospect
  Theory Framework}.
\newblock \bibinfo{journal}{\emph{IEEE Trans. Automat. Control}}
  \bibinfo{volume}{63} (\bibinfo{year}{2018}), \bibinfo{pages}{2867--2882}.
\newblock


\bibitem[\protect\citeauthoryear{Jie, Xu, Wang, Wang, and Shen}{Jie
  et~al\mbox{.}}{2021}]%
        {Jie2021BiddingVC}
\bibfield{author}{\bibinfo{person}{Cheng Jie}, \bibinfo{person}{Da Xu},
  \bibinfo{person}{Zigeng Wang}, \bibinfo{person}{Lu Wang}, {and}
  \bibinfo{person}{Wei-Yuan Shen}.} \bibinfo{year}{2021}\natexlab{}.
\newblock \showarticletitle{Bidding via Clustering Ads Intentions: an Efficient
  Search Engine Marketing System for E-commerce}.
\newblock \bibinfo{journal}{\emph{ArXiv}}  \bibinfo{volume}{abs/2106.12700}
  (\bibinfo{year}{2021}).
\newblock


\bibitem[\protect\citeauthoryear{Killoran}{Killoran}{2013a}]%
        {Killoran2013HowTU}
\bibfield{author}{\bibinfo{person}{J. Killoran}.}
  \bibinfo{year}{2013}\natexlab{a}.
\newblock \showarticletitle{How to Use Search Engine Optimization Techniques to
  Increase Website Visibility}.
\newblock \bibinfo{journal}{\emph{IEEE Transactions on Professional
  Communication}}  \bibinfo{volume}{56} (\bibinfo{year}{2013}),
  \bibinfo{pages}{50--66}.
\newblock


\bibitem[\protect\citeauthoryear{Killoran}{Killoran}{2013b}]%
        {6463486}
\bibfield{author}{\bibinfo{person}{John~B. Killoran}.}
  \bibinfo{year}{2013}\natexlab{b}.
\newblock \showarticletitle{How to Use Search Engine Optimization Techniques to
  Increase Website Visibility}.
\newblock \bibinfo{journal}{\emph{IEEE Transactions on Professional
  Communication}} \bibinfo{volume}{56}, \bibinfo{number}{1}
  (\bibinfo{year}{2013}), \bibinfo{pages}{50--66}.
\newblock
\urldef\tempurl%
\url{https://doi.org/10.1109/TPC.2012.2237255}
\showDOI{\tempurl}


\bibitem[\protect\citeauthoryear{Li, Song, Sun, and Zhu}{Li
  et~al\mbox{.}}{2021}]%
        {li2021frequentnet}
\bibfield{author}{\bibinfo{person}{Yifei Li}, \bibinfo{person}{Kuangyan Song},
  \bibinfo{person}{Yiming Sun}, {and} \bibinfo{person}{Liao Zhu}.}
  \bibinfo{year}{2021}\natexlab{}.
\newblock \showarticletitle{FrequentNet: A Novel Interpretable Deep Learning
  Model for Image Classification}.
\newblock \bibinfo{journal}{\emph{Available at SSRN:
  https://ssrn.com/abstract=3895462}} (\bibinfo{year}{2021}).
\newblock


\bibitem[\protect\citeauthoryear{Lin, Jie, and Marcus}{Lin
  et~al\mbox{.}}{2018}]%
        {LIN20181}
\bibfield{author}{\bibinfo{person}{Kun Lin}, \bibinfo{person}{Cheng Jie}, {and}
  \bibinfo{person}{Steven~I. Marcus}.} \bibinfo{year}{2018}\natexlab{}.
\newblock \showarticletitle{Probabilistically distorted risk-sensitive
  infinite-horizon dynamic programming}.
\newblock \bibinfo{journal}{\emph{Automatica}}  \bibinfo{volume}{97}
  (\bibinfo{year}{2018}), \bibinfo{pages}{1--6}.
\newblock
\showISSN{0005-1098}
\urldef\tempurl%
\url{https://doi.org/10.1016/j.automatica.2018.07.028}
\showDOI{\tempurl}


\bibitem[\protect\citeauthoryear{Mikolov, Sutskever, Chen, Corrado, and
  Dean}{Mikolov et~al\mbox{.}}{2013}]%
        {Mikolov2013DistributedRO}
\bibfield{author}{\bibinfo{person}{Tomas Mikolov}, \bibinfo{person}{Ilya
  Sutskever}, \bibinfo{person}{Kai Chen}, \bibinfo{person}{G. Corrado}, {and}
  \bibinfo{person}{J. Dean}.} \bibinfo{year}{2013}\natexlab{}.
\newblock \showarticletitle{Distributed Representations of Words and Phrases
  and their Compositionality}. In \bibinfo{booktitle}{\emph{NIPS}}.
\newblock


\bibitem[\protect\citeauthoryear{PrashanthL., Jie, Fu, Marcus, and
  Szepesvari}{PrashanthL. et~al\mbox{.}}{2016}]%
        {PrashanthL2016CumulativePT}
\bibfield{author}{\bibinfo{person}{A. PrashanthL.}, \bibinfo{person}{Cheng
  Jie}, \bibinfo{person}{M. Fu}, \bibinfo{person}{S. Marcus}, {and}
  \bibinfo{person}{Csaba Szepesvari}.} \bibinfo{year}{2016}\natexlab{}.
\newblock \showarticletitle{Cumulative Prospect Theory Meets Reinforcement
  Learning: Prediction and Control}. In \bibinfo{booktitle}{\emph{ICML}}.
\newblock


\bibitem[\protect\citeauthoryear{Yalcin and Kose}{Yalcin and Kose}{2010}]%
        {Yalcin2010WhatIS}
\bibfield{author}{\bibinfo{person}{Nursel Yalcin} {and} \bibinfo{person}{Utku
  Kose}.} \bibinfo{year}{2010}\natexlab{}.
\newblock \showarticletitle{What is search engine optimization: SEO?}
\newblock \bibinfo{journal}{\emph{Procedia - Social and Behavioral Sciences}}
  \bibinfo{volume}{9} (\bibinfo{year}{2010}), \bibinfo{pages}{487--493}.
\newblock


\bibitem[\protect\citeauthoryear{Zhang and Cabage}{Zhang and Cabage}{2017}]%
        {Zhang2017SearchEO}
\bibfield{author}{\bibinfo{person}{Sonya Zhang} {and} \bibinfo{person}{Neal
  Cabage}.} \bibinfo{year}{2017}\natexlab{}.
\newblock \showarticletitle{Search Engine Optimization: Comparison of Link
  Building and Social Sharing}.
\newblock \bibinfo{journal}{\emph{Journal of Computer Information Systems}}
  \bibinfo{volume}{57} (\bibinfo{year}{2017}), \bibinfo{pages}{148 -- 159}.
\newblock


\bibitem[\protect\citeauthoryear{Zhao, Zhan, and Jie}{Zhao
  et~al\mbox{.}}{2018}]%
        {ZHAO2018619}
\bibfield{author}{\bibinfo{person}{Xinyan Zhao}, \bibinfo{person}{Mengqi Zhan},
  {and} \bibinfo{person}{Cheng Jie}.} \bibinfo{year}{2018}\natexlab{}.
\newblock \showarticletitle{Examining multiplicity and dynamics of publics’
  crisis narratives with large-scale Twitter data}.
\newblock \bibinfo{journal}{\emph{Public Relations Review}}
  \bibinfo{volume}{44}, \bibinfo{number}{4} (\bibinfo{year}{2018}),
  \bibinfo{pages}{619--632}.
\newblock
\showISSN{0363-8111}
\urldef\tempurl%
\url{https://doi.org/10.1016/j.pubrev.2018.07.004}
\showDOI{\tempurl}


\bibitem[\protect\citeauthoryear{Zhu, Jarrow, and Wells}{Zhu
  et~al\mbox{.}}{2021a}]%
        {zhu2021time}
\bibfield{author}{\bibinfo{person}{Liao Zhu}, \bibinfo{person}{Robert~A.
  Jarrow}, {and} \bibinfo{person}{Martin~T. Wells}.}
  \bibinfo{year}{2021}\natexlab{a}.
\newblock \showarticletitle{Time-Invariance Coefficients Tests with the
  Adaptive Multi-Factor Model}.
\newblock \bibinfo{journal}{\emph{arXiv preprint arXiv:2011.04171}}
  (\bibinfo{year}{2021}).
\newblock


\bibitem[\protect\citeauthoryear{Zhu, Sun, and Wells}{Zhu
  et~al\mbox{.}}{2022}]%
        {zhu2022clustering}
\bibfield{author}{\bibinfo{person}{Liao Zhu}, \bibinfo{person}{Ningning Sun},
  {and} \bibinfo{person}{Martin~T. Wells}.} \bibinfo{year}{2022}\natexlab{}.
\newblock \showarticletitle{Clustering Structure of Microstructure Measures}.
\newblock \bibinfo{journal}{\emph{Applied Economics and Finance}}
  \bibinfo{volume}{9}, \bibinfo{number}{1} (\bibinfo{year}{2022}),
  \bibinfo{pages}{85--95}.
\newblock


\bibitem[\protect\citeauthoryear{Zhu, Wu, and Wells}{Zhu
  et~al\mbox{.}}{2021b}]%
        {zhu2021news}
\bibfield{author}{\bibinfo{person}{Liao Zhu}, \bibinfo{person}{Haoxuan Wu},
  {and} \bibinfo{person}{Martin~T. Wells}.} \bibinfo{year}{2021}\natexlab{b}.
\newblock \showarticletitle{A News-based Machine Learning Model for Adaptive
  Asset Pricing}.
\newblock \bibinfo{journal}{\emph{arXiv preprint arXiv:2106.07103}}
  (\bibinfo{year}{2021}).
\newblock


\end{thebibliography}

\end{document}